\documentclass[11pt]{article} 

\usepackage{amsmath,amssymb}
\usepackage{multicol}
\usepackage{fullpage}
\usepackage{graphicx}
\usepackage{svg}
\usepackage{comment}
\usepackage{bbm}
\usepackage{subcaption}
\usepackage{natbib}
\usepackage{mathtools}

\newtheorem{theorem}{Theorem}[section]

\newtheorem{proposition}[theorem]{Proposition}

\usepackage{xcolor}
\usepackage{stmaryrd}
\usepackage{enumitem}
\usepackage{xspace}
\usepackage{hyperref}
\usepackage{tikz}
\usepackage{multirow}
\usepackage{booktabs}
\usepackage{algorithm}
\usepackage{algorithmic}

\newcommand{\code}[1]{{\texttt{#1}}}

\newcommand{\R}{\mathbb{R}}

\newcommand{\E}{\mathbb{E}}
\newcommand{\A}{\mathcal{A}}
\newcommand{\D}{\mathcal{D}}

\newcommand{\M}{\mathcal{M}}

\renewcommand{\S}{\mathcal{S}}

\newcommand{\wait}{\code{wait}}
\renewcommand{\stop}{\code{stop}}
\newcommand{\step}{\code{step}}

\newcommand{\vcentered}[1]{\begin{tabular}{l} #1 \end{tabular}}

\usepackage{natbib}
 \bibpunct[, ]{(}{)}{,}{a}{}{,}%
\title{Learning Algorithms for Regenerative Stopping Problems with Applications to Shipping Consolidation in Logistics}
\author{
\begin{tabular}{cc}
\begin{tabular}{c}
Kishor Jothimurugan\thanks{Work done while the author was at Nokia Bell Labs.}\\
University of Pennsylvania
\end{tabular}\qquad
&
\qquad
\begin{tabular}{c}
Matthew Andrews\\Nokia Bell Labs
\end{tabular}\\ 
\\
\begin{tabular}{c}
Jeongran Lee\\Nokia Bell Labs
\end{tabular}\qquad
&
\qquad
\begin{tabular}{c}
Lorenzo Maggi\\Nokia Bell Labs
\end{tabular}
\end{tabular}
}

\date{}

\begin{document}

\maketitle

\begin{abstract}
We study regenerative stopping problems in which the system starts anew whenever the controller decides to stop and the long-term average cost is to be minimized. Traditional model-based solutions involve estimating the underlying process from data and computing strategies for the estimated model. In this paper, we compare such solutions to deep reinforcement learning and imitation learning which involve learning a neural network policy from simulations. We evaluate the different approaches on a real-world problem of shipping consolidation in logistics and demonstrate that deep learning can be effectively used to solve such problems.
\end{abstract}



%

\section{Introduction}\label{sec:intro}
Artificial Intelligence (AI) techniques such as Reinforcement Learning (RL) and Imitation Learning (IL) are adept at making decisions in complex environments such as video games and robot navigation. In this work we investigate whether these techniques can provide efficient solutions for problems that arise in Operations Research. 


We focus on a classic regenerative stopping problem where, as long as a stopping decision is not made, costs are accumulated over time and the state continues to evolve. 
Both costs and state evolution are governed by a stochastic data arrival process. When the controller decides to stop, then an immediate cost is incurred and the state is reset to the start state. The goal is to minimize the long-run average cost.

A prominent instance of this class of stopping problems arises in logistics. We consider a transportation hub that sends truckloads of goods to different destinations. There is a cost for shipping a truck to a destination and the goods have delay requirements on their delivery. Our goal is to decide when to send a truck to a destination. If we delay sending a truck then more goods might arrive for the destination, meaning that we can better {\em consolidate} goods and send fewer trucks. 
On the other hand, waiting may cause disruptions to the end-to-end delivery process. Our goal is to manage this trade-off according to a cost function that includes both shipping and delay costs. The stochastic input data process represents orders being placed by a customer, while the state describes the orders waiting to be shipped.

Assuming that the statistical properties of the input arrival process are unknown, in this work we study the efficacy of learning-based solutions for regenerative stopping problems. In particular we compare the following three techniques.

\begin{itemize}
    \item \textit{Model-based approach}. Here we use the Markov Decision Process (MDP) solution based on an estimate of the model parameters. Specifically, future input arrival statistics are predicted, the associated approximate MDP is solved, the solution is applied for a few steps, and the process is repeated. As time goes by, the parameter estimate improves and hence the solution quality does too.
    \item \textit{Deep reinforcement learning} (DRL). Here we run a deep value-based or policy-based RL algorithm such as Deep Q-Network (DQN) \citep{mnih2015human} or Proximal Policy Optimization (PPO) \citep{schulman2017proximal}. These algorithms learn the optimal action for any given state without explicitly learning the model. The term ``deep'' implies that a neural network is used to approximate the value function and/or the policy.
    \item \textit{Imitation learning} (IL). Here we run a hindsight optimization algorithm for each time step in the past to determine what the best decision would have been assuming that we know the future. We then apply the Imitation Learning algorithm from \cite{ross2011reduction} to imitate this hindsight optimal solution in real-time. 
\end{itemize}

We will show that the specific nature of the problem at hand makes it appealing to utilize \emph{ad-hoc} techniques such as the model-based approach and IL, thanks to the efficiency of the respective solutions. Indeed, the MDP in the model-based approach can be efficiently solved via an iterative technique involving finding the root of a Lagrangian-type function, as studied by \cite{miller1981countable}. 
On the other hand, in our scenario the hindsight optimal solution in IL can be solved in polynomial time. 

Our experiments will show that, by directly learning the optimal actions from the input data without constructing an explicit prediction for future inputs, one can better adapt to changes in the input distribution. Two key questions need to be addressed for each of the investigated policy learning approaches (DRL and IL). First, we need to define the state space in a way that is rich enough to enable learning. Second, most learning algorithms require a significant amount of training data. For most practical problems a single pass through the available data may not be sufficient to learn a good policy. We therefore present ways in which we can reuse the available training data via multiple passes.

In the remainder of the paper we define each approach in detail and compare their performance. As mentioned, we shall compare these techniques on a shipping consolidation problem that arises in transportation logistics. However, our techniques apply to a more generic problem involving optimal stopping on sequential data. Hence we begin by describing our abstract stopping problem and then present the shipping consolidation problem as a special case.

\subsection*{Related work}
Traditional optimization methods have been applied to a variety of shipping consolidation problems, e.g.\ 
\cite{UsterA10}, \cite{Deng2013} and \cite{Zhang15}.
Our study falls into a line of work that uses Machine Learning (ML) techniques to solve optimization problems. For example, \cite{kong2018a} applies RL to knapsack and secretary problems, and \cite{khalil2017learning} uses RL to solve graph problems. \cite{mazyavkina2020reinforcement} and \cite{bengio2020machine} provide an extensive survey on applications of ML and RL in combinatorial optimization. 
For the specific problem of shipping optimization, \cite{vanAndel18} uses ML to consolidate shipments from nearby suppliers. 

Classic stopping problems prescribe that a system terminates upon a stopping decision. Traditional optimization methods for such problems are presented in  \cite{DesaiFC12,Belomestny13}.
Some RL solutions for stopping problems exist in the literature, e.g., by \cite{tsitsiklis1999optimal, yu2007q, choi2006generalized}. Also, \cite{becker2019deep} apply deep learning to stopping problems that arise in financial option pricing. However, to the best of our knowledge we are the first to apply RL to \emph{regenerative} stopping problems, where the system starts anew after a stop and the cost is minimized in the long run.

\section{The Model}\label{sec:model}
We consider a continuous-time MDP under average reward criterion with two fundamental constraints: (i) the action space is binary ($\wait$/$\stop$) and (ii) there is a state $s_0$ such that, whenever the \stop\ action is chosen, then the underlying system state resets to $s_0$.

More specifically, the model of the system is given by a tuple $\M = (\S,\A,X,\delta,\eta,s_0, c_{\wait}, c_{\stop})$ where $\mathcal S$ is a countable set of states. In each state $s\in \mathcal S$, a set of actions $\A(s)\subseteq \{\wait, \stop\}$ is available to the decision making agent. The initial state is $s_0 \in \S$ and $\A(s_0)=\{\wait\}$. $X\subseteq \R^n$ denotes the set of \emph{data points} that govern the transition between states and $\delta:\S\times X\to\S$ is the (deterministic) transition function. $\eta: X\times\R_{\geq 0}\to [0,1]$ is the distribution governing the data arrival process---i.e., $\eta(x,\tau)$ denotes the probability that data point $x$ arrives after time $\tau$. $c_{\wait}, c_{\stop}: \S\to \R$ define the cost per unit of time and the lump cost for taking the actions $\wait$ and $\stop$, respectively.

The process starts at time $t=0$ in state $s_0$. Suppose that at time $t>0$ the system is in state $s$. If the $\stop$ action is taken, then an instantaneous cost $c_{\stop}(s)$ is incurred and the system immediately resets its state $s:=s_0$. If the $\wait$ action is taken, then the state has no immediate transition and a cost is accumulated at a rate of $c_{\wait}(s)$ per unit time. In either case, at time $t+\tau$ a new data point $x\in X$ materializes and drives the next state transition, where $\tau$ is a random time interval. More specifically, at time $t+\tau$ the system jumps to state $s':=\delta(s,x)$. 
We also denote by $\delta(s,x,a)$ the state transition on action $a$ which is given by $\delta(s,x,\stop) = \delta(s_0,x)$ and $\delta(s,x,\wait) = \delta(s,x)$. 
By assuming that the process of data arrival $(x,\tau)$ is \emph{i.i.d.}, (i) the resulting transition kernel is Markovian with probability $p(s'|s,a)=\Pr(\delta(s,x,a)=s')$ --- where the probability is with respect to the random input $x$ --- and (ii) the state transition process with inter-arrival time $\tau$ becomes a renewal stochastic process. 
For the simplicity of analysis we will also assume $x,\tau$ to be independent.

A \emph{policy} $\pi:\mathcal S\rightarrow[0,1]$ maps a state $s\in\S$ to the probability $\pi(s)$ of choosing the $\stop$ action. Let us denote by $T_{\stop}^i$ the time at which the $\stop$ action was chosen for the $i^{\text{th}}$ time. We can express the cost accumulated between two consecutive times when the $\stop$ action is selected as
\[
c^{\pi}(T_{\stop}^i,T_{\stop}^{i+1}) := c_{\stop}(s(T_{\stop}^{i+1})) + \int_{T_{\stop}^i}^{T_{\stop}^{i+1}} c_{\wait}(s(t)) \, dt
\]
where $s(t)$ is the state at time $t$. Similarly, we can define $C^{\pi}(T)$ as the total cost accumulated up to time $T>0$, interpreted as a random variable whose realizations are given by a function of the observed data points and the actions chosen by $\pi$. More precisely, if $T_{\stop}^j$ is the last time before $T$ at which the $\stop$ action was chosen, then
\[
C^{\pi}(T) := \sum_{i=1}^j c^{\pi}(T_{\stop}^{i-1},T_{\stop}^i) + \int_{T_{\stop}^j}^{T} c_{\wait}(s(t)) \, dt. 
\]
where by convention $T_{\stop}^0:=0$.

In the context of learning, we assume that the distribution $\eta$ is \emph{unknown} to the learner. On the other hand, the learner has full knowledge of the transition function $\delta$. Then, our goal is to learn a policy $\pi^*$ minimizing the long-term average cost given by
\begin{align}
J(\pi) &= \limsup_{T\rightarrow\infty} \, \mathbb E_{\eta} \, \left[\frac{C^{\pi}(T)}{T}\right] \notag \\
&= \limsup_{j\rightarrow\infty} \, \mathbb E_{\eta} \, \left[\frac{\sum_{i=1}^j c^{\pi}(T_{\stop}^{i-1},T_{\stop}^i) }{T_{\stop}^j}\right]. \label{eq:objective}
\end{align}

\subsection{An Application to Supply-chain: Shipping Consolidation}

We now show how the general model above can describe an important problem in supply chain optimization that we refer to as {\em shipping consolidation}. At a warehouse, orders arrive sequentially and are to be shipped to customers and we focus on all those orders that have to be shipped to one specific destination. Upon the arrival of each new order, the decision is on whether to consolidate and ship all the orders that are currently held at the warehouse (the $\stop$ action), or to wait for more orders (the $\wait$ action). Since shipping fees are typically an increasing concave function of the weight of the shipment, we should only ship when the total weight of orders is close to truck capacity if the goal is simply to minimize shipping cost. However, there is a clear incentive in not letting the customer wait too long, which leads to a shipping cost-vs-delay trade-off that we aim to resolve.

We denote by $\ell$ the truck load, namely the sum of weights of orders waiting to be shipped. We let $n$ be the number of items currently waiting to be shipped, and we define the state of the system by the pair $s:=(\ell,n)$. The observed \emph{data point} $x\in X\subseteq\R_{\geq 0}$ is the weight of the new order. Hence, the state transition law can be defined simply as $\delta(s,x)=(\ell+x,n+1)$. The distribution $\eta$ is given by $\eta(x,\tau) = P_{X}(x)P_{I}(\tau)$ where $P_{X}$ and $P_{I}$ are the probabilities over the possible weights and inter-arrival times of the next order, respectively.

Let $L$ be the truck load capacity. If we assume that the weights of the individual orders are much smaller than the capacity $L$ ($x\ll L, \ \forall\, x\in X$) then we can suppose that \emph{all} waiting orders can be shipped when consolidation occurs. Therefore, upon consolidation the system always resumes from the empty truck state $s_0=(\ell=0,n=0)$. Furthermore, when the truck is full, only the \stop{} action is available, i.e., $\mathcal A(s)=\{\stop\}$ for all states $s=(\ell,n)$ with $\ell\ge L$.

We let $f(\ell)$ be the shipping cost incurred if a truck with load $\ell$ is shipped. We reasonably assume $f(.)$ to be increasing and concave with respect to the truck load $\ell$. The cost of \emph{consolidating} a shipment in state $s$ is the instantaneous cost $c_{\stop}(\ell, n)=f(\ell)$.
We quantify the delay cost as the time spent by an order waiting at the warehouse, summed over all orders currently being held off. For this purpose, we define the rate at which cost is accumulated over time when the \wait{} action is chosen in state $s$ as $c_{\wait}(\ell,n)=\alpha n$, where the parameter $\alpha\ge 0$ regulates the trade-off between shipping cost and delay.

We remark that for many real-world shipping hubs we will have a {\em set} of shipping problems, each of which corresponds to a separate destination and has specific order arrival process, shipping fee $f(\ell)$ and sensitivity parameter $\alpha$. Although each problem is independent and can be solved separately, in our learning-based solutions we will use findings for one shipping destination to help guide the solution for another. 

\section{A Model-based Solution}\label{sec:model_based}
We observe that the continuous-time Markov Decision Process (MDP) underlying our regenerative stopping problem is fully characterized by the distribution $\eta$ of inter-arrival times $\tau$ and data points $x\in X$. Thus, a natural attempt to learn the optimal strategy is, in a classic model-based fashion, to (i) predict the future distribution $\eta$ as $\widehat \eta$, 
(ii) solve the MDP associated with $\widehat \eta$ and (iii) apply the MDP solution for the next (few) step(s) and then return to (i). The model-based algorithm is shown in Algorithm~\ref{alg:model}.

It is worthwhile to describe how the special structure of the problem at hand can help us to come up with a specialized, low complexity algorithm for solving the underlying MDP in step (ii) above. 
For a detailed analysis we refer to the work by \cite{miller1981countable}.

\begin{algorithm}[t]
\begin{algorithmic}[1]
\FOR{data points arriving at time $t_m$, $m\in \mathbb N$}
\STATE Predict the future input data and inter-arrival distribution $\eta$ as $\widehat{\eta}$
\STATE Compute $\pi_{\hat{\eta}}\leftarrow\code{MDP}(\widehat \eta)$ 
\STATE Apply the action $\pi_{\hat{\eta}}(s(t_m))$
\ENDFOR

\FUNCTION{$\code{MDP}$($\widehat \eta$)}
\STATE Find the root $\nu^*$ of the function $J_{\widehat \eta}:\R_{\geq 0}\rightarrow \mathbb R$ computed via \texttt{relaxed-stopping}($\nu,\widehat \eta$)
\STATE \textbf{return} the stopping policy $\pi^*:=\pi^*_{\nu^*}$
\ENDFUNCTION

\FUNCTION{\texttt{relaxed-stopping}($\nu,\widehat \eta$)}
\STATE Compute via dynamic programming $J_{\widehat \eta}(\nu)=\min_{\pi} \mathbb{E}_{\widehat \eta} \left[ c^{\pi}(0,T^1_{\stop}) - \nu T^1_{\stop}\right]$ and the associated optimal policy $\pi^*_{\nu}$. \\
State, action, and cost definitions are as in Section \ref{sec:model}, expect for the instantaneous cost for state/action pair $(s,\wait)$ being $\mathbb E_{\widehat \eta}[\tau] c_{\wait}(s)$ and for the state transition law distributed according to the estimated $\widehat \eta$.
\STATE \textbf{return} $J_{\widehat \eta}(\nu)$ and $\pi^*_{\nu}$
\ENDFUNCTION

\end{algorithmic}
\caption{Model-based algorithm}
\label{alg:model}
\end{algorithm}

We first observe that, the input data $X$ and inter-arrival times $\tau$ being \emph{i.i.d.}, a stopping strategy $\pi$ engenders a random sequence of stopping times $T_{\stop}^1,T_{\stop}^2,\dots$ constituting a renewal process. In other words, the inter-stopping times $T_{\stop}^i-T_{\stop}^{i-1}$ are themselves \emph{i.i.d.}. 

Then, we can rewrite the minimum long-term average cost in \eqref{eq:objective} as a sole function of the \emph{first} stopping time as follows:
\begin{align}
\inf_{\pi}\, J(\pi) &=  \inf_{\pi}\limsup_{T\rightarrow\infty} \, \mathbb E_{\eta} \, \left[\frac{C^{\pi}(T)}{T}\right] \\ &= \inf_{\pi} \frac{\mathbb{E}_{\eta}[c^{\pi}(0,T^1_{\stop})]}{\mathbb{E}_{\eta}\left[T_{\stop}^1\right]} \label{eq:from_renewal} \\
&= \inf_{\nu\ge 0} : \left\{ \min_{\pi} \mathbb{E}_{\eta} \left[ c^{\pi}(0,T^1_{\stop}) - \nu T^1_{\stop} \right]<0 \right\} \label{eq:rewrite}
\end{align}
where the Equation~(\ref{eq:from_renewal}) stems from classic renewal theory (e.g., \cite{cox1962renewal}) and claims that the average reward in the long-run equals the ratio between the average cost accumulated until the first stopping action and the average first stopping time. Equation \eqref{eq:rewrite} holds since the function $J_{\eta}(\nu):=\min_{\pi} \mathbb{E}_{\eta} \left[ c^{\pi}(0,T^1_{\stop}) - \nu T^1_{\stop}\right]$ has a single root, that coincides with the optimal value. We note that we can interpret $\nu$ as a penalty for waiting $T^1_{\stop}$ units of time before stopping. 

From \eqref{eq:rewrite} it stems that the optimization can be then conveniently split into two tasks: 
(i) computing $J_{\eta}(\nu)$ and the associated optimal strategy as a sub-routine and (ii) finding the root $\nu^*$ of the function $J_{\eta}$. Task (ii) can be accomplished using, e.g., a binary search for $\nu^*$. Task (i) can be solved via dynamic programming, after turning the continuous-time problem at hand into a discrete-time one by naturally assigning the instantaneous (average) waiting cost $\overline{c}_{\wait}(s)=\mathbb E_{(x,\tau)\sim\eta}[\tau] \cdot c_{\wait}(s)$ if the \wait{} action is chosen in the state $s$.

\section{Learning from Data}\label{sec:learning}
The downside of a model-based approach as described in the previous section is that it fully delegates to the prediction module (Algorithm \ref{alg:model}, line 2) the task of figuring out how the input data will change over time. The main goal of this paper is to evaluate techniques that try to directly learn the optimal actions from the input data without actually constructing an explicit prediction for future inputs. We will see in Section~\ref{sec:exp} that such techniques can actually outperform model-based techniques since the former can quickly adapt when the input parameters change.  

We now describe two types of deep learning based solutions that learn a policy $\tilde{\pi}$ from historical data. We assume that we have data collected over a fixed period of time $T$. Let $\Gamma = \{\rho_1,\ldots,\rho_N\}$ denote the dataset where each $\rho_i = (x_{i,1},\tau_{i,1}),\ldots,(x_{i,m_i},\tau_{i,m_i})$ is a finite sequence of (timed) data points with $x_{i,j}\in X$ and $\tau_{i,j}\in\R_{\ge 0}$. We have for all $i \in \{1,\ldots,N\}$, $\sum_{j=1}^{m_i}\tau_{i,j} \leq T$. We denote by $C^{\pi}_\rho(T)$ the expected total cost accumulated up to time $T$, when sampling actions from policy $\pi$ and the sequence of data points is fixed to $\rho$. We denote by $\D$ the distribution over finite sequences of timed data points in $X\times\R_{\geq 0}$ where the data points are sampled \emph{i.i.d.} from $\eta$ until the total time exceeds $T$ and the last data point is omitted. Then, the learning objective is given by
\begin{align}\label{eqn:learning}
    \underset{\pi\in\Pi}{\operatorname{\text{minimize}}}\,  \frac{1}{T}\E_{\rho\sim\D}[C^{\pi}_\rho(T)],
\end{align}
where $\Pi$ is a class of policies. Note that the objective here is a finite horizon approximation of the objective defined in Section~\ref{sec:model}. Since each sequence $\rho_i\in\Gamma$ is drawn \emph{i.i.d.} from the distribution $\D$, we minimize the empirical cost---i.e., we solve
\begin{align*}
    \underset{\pi\in\Pi}{\operatorname{\text{minimize}}} \, \frac{1}{NT}\sum_{i=1}^NC^{\pi}_{\rho_i}(T).
\end{align*}

\subsection{Deep Reinforcement Learning}\label{sec:drl}
Deep Reinforcement Learning (DRL) has been successful in solving challenging problems in many domains such as game playing \citep{mnih2015human, silver2016mastering} and robot control \citep{silver2014deterministic, collins2005efficient}. In DRL the underlying system is modeled as a discrete time MDP with unknown dynamics and the policy is represented using a neural network. DRL algorithms seek to learn the parameters of the neural network in order to maximize the expected reward (alternatively, minimize the expected cost).

Formally, a discrete-time MDP is given by $\tilde{\M} = (\tilde{\S}, \tilde{\A}, P, R, \gamma, \eta_0)$ where $\tilde{\S}$ is the set of states, $\tilde{\A}$ is the set of actions, $P(s'| s,a)$ is the probability of transitioning to $s'$ when action $a$ is taken in state $s$, $R(s,a, s')$ is the reward for taking action $a$ in state $s$ and transitioning to $s'$, $\gamma$ is the discount factor and $\eta_0$ is the initial state distribution. A policy is a function $\tilde{\pi}: \tilde{\S}\times \tilde{\A}\to [0,1]$ which defines the probability $\tilde{\pi}(a| s)$ of taking action $a$ in state $s$. The value of a policy $\tilde{\pi}$ at state $s$ is the expected reward starting at $s$ given by 
$$V^{\tilde{\pi}}(s) = \E\Big[\sum_{t=0}^\infty \gamma^t R(s_t, a_t, s_{t+1})\mid s_0 = s, \tilde{\pi}\Big].$$

Reinforcement learning algorithms aim to compute a policy $\tilde{\pi}^* = \operatorname*{\arg\max}_{\tilde{\pi}}\E_{s\sim\eta_0}[V^{\tilde{\pi}}(s)]$. Typically, these algorithms treat the MDP as a black box and only require the ability to sample episodes (sequences of transitions in the MDP) using any policy $\tilde{\pi}$. 

Given a continuous-time system model $\M$ and a time-limit $T_d$, we can define a discrete-time MDP $\tilde{\M}$ with state space $\tilde{\S} = (\S\times[0,T_d])$ and action space $\tilde{\A} = \A$. A state $(s,t)\in\tilde{\S}$ denotes that the system transitioned to $s$ at time $t$. The transition probabilities and rewards are defined in a natural way so that the following proposition holds.

\begin{proposition}
Given a system model $\M$ and a time-limit $T_d$, we can define a discrete-time MDP $\tilde{\M}$ such that for any policy $\tilde{\pi}$ for $\tilde{\M}$ we can construct a policy $\pi$ for $\M$ (and vice-versa) such that
$$
\E_{(s,t)\sim\eta_0}[V^{\tilde{\pi}}(s,t)] = -\E_{\rho\sim\D}[C_{\rho}^{\pi}(T_d)].
$$
\end{proposition}

Therefore, we can use existing DRL algorithms to learn a policy that minimizes the finite time approximation of the objective in (\ref{eq:objective}).


\begin{algorithm}[t]
\begin{algorithmic}[1]
\FUNCTION{GetEpisode($\tilde{\pi}$, $\Gamma$, $T_d$)}
\STATE Initialize $\code{ep} \leftarrow []$; $s\leftarrow s_0$; $t\leftarrow 0$
\STATE Select $\rho\sim \code{Uniform}(\Gamma)$
\STATE Select $T_s\sim \code{Uniform}([0, T-T_d])$
\FOR{$(x_i,\tau_i)$ in $\rho[T_s:T_s+T_d]$}
\STATE Sample action $a\sim\tilde{\pi}(\cdot\mid (s,t))$
\STATE $s' \leftarrow \delta(s, x_i, a)$
\STATE $t'\leftarrow t+\tau_i$
\IF{a = $\wait$}
\STATE $r \leftarrow -c_{\wait}(s)\tau_i$
\ELSE
\STATE $r \leftarrow -c_{\wait}(s_0)\tau_i-c_{\stop}(s)$
\ENDIF
\STATE Add $((s,t),a,r,(s',t'))$ to $\code{ep}$
\STATE $s\leftarrow s'$; $t\leftarrow t'$
\ENDFOR
\STATE \textbf{return} $\code{ep}$
\ENDFUNCTION
\end{algorithmic}
\caption{Sample an episode from $\tilde{\M}$}
\label{alg:episodes}
\end{algorithm}

Algorithm~\ref{alg:episodes} describes the process of collecting episodes from $\tilde{\M}$. To sample an episode, we first choose a sequence $\rho$ uniformly at random from $\Gamma$. We then pick a starting time $T_s$ from the interval $[0,T-T_d]$ uniformly at random. We then use $\rho[T_s:T_s+T_d]$ to simulate $\tilde{\M}$ where $\rho[T_s:T_s+T_d]$ is the contiguous subsequence of $\rho$ consisting of data items that arrived after time $T_s$ and before time $T_s+T_d$. It is usually the case that historic data is available for a long duration but the number of data sequences is small (typically $|\Gamma|=1$). In such cases, this heuristic generates different scenarios for the RL agent to train on, thereby improving generalization.

\subsection{Imitation Learning}
Imitation learning techniques aim to mimic an \emph{expert} in a given task. Typically, the expert is a human who provides the best actions to take in some finite number of states. Then, imitation learning methods use supervised learning to learn a mapping (e.g., a neural network) from states to actions. In our case, we do not require a human expert --- instead, we can leverage the structure of the problem to compute the best actions using \emph{hindsight optimization} as described below. \\

\textbf{Hindsight Optimization.} Given a sequence of data points collected during a certain time interval, we now show how to compute the best sequence of actions minimizing the sum of costs over that time interval, assuming that all data points are known in advance. More precisely, we are given a sequence of timed data points $\rho = (x_1,\tau_1),\ldots,(x_m,\tau_m)$ with $x_i\in X$ and $\tau_i\in\R_{\geq 0}$, and we solve for the best sequence of actions $a_0^*,\ldots,a_{m}^*$ that minimize the cost for this particular sequence of data points. We will prove that this can be computed efficiently.

Let us denote by $Q^{\rho}_j: \S\times \A \to \R$ the minimum cost function, i.e., $Q^{\rho}_j(s, a)$ denotes the minimum cost that can be achieved if the current is $s$, current action is $a$, current time is $t_j = \sum_{k=1}^{j}\tau_k$ and the next data item is $x_{j+1}$. For $s\in\S$, $a\in\A$ and $t\in\R_{\geq 0}$ we denote by $c_{\step}(s,a,t)$ the cost for taking action $a$ in state $s$ and waiting for $t$ time units---i.e., $c_{\step}(s,\stop,t) = c_{\stop}(s) + c_{\wait}(s_0)t$ and $c_{\step}(s,\wait,t) = c_{\wait}(s)t$. Now $Q^{\rho}_j$ can be computed recursively as follows.

\noindent \textit{Base Case.} For $j = m$, the minimum cost is given by $Q^{\rho}_{m}(s,a) = c_{\step}(s,a,T_d-t_m)$ where $T_d$ is the time duration for which the data in $\rho$ was collected.

\noindent \textit{Inductive Case.} For $j < m$, the minimum cost is
    \begin{align*}
        Q^{\rho}_j(s,a) = c_{\step}(s,a,\tau_{j+1}) + V^{\rho}_{j+1}(\delta(s,x_{j+1},a)),
    \end{align*}
    where $V^{\rho}_{j+1}(s') = \min_{a\in \A}Q^{\rho}_{j+1}(s',a)$ for all $s'\in\S$.

Once we have the $Q$ values, the optimal actions are the actions that minimize the $Q$ value for the state visited in the particular time-step. More precisely,
\begin{align*}
a_j^* = & \, \operatorname*{\arg\min}_{a\in \A}Q^{\rho}_j(s_j, a), \ \ \forall j\geq0
\end{align*}
where $s_j = \delta(s_{j-1}, x_j, a_{j-1}^*)$. 
Regarding the algorithm complexity, we observe that at iteration $j$ there are at most $j+1$ possible states since the $\stop$ action always resets to $s_0$. Therefore, in iteration $j$ the algorithm only performs $O(j+1)$ operations. The following proposition follows.
\begin{proposition} \label{prop:complexity_hindsight}
The hindsight optimization finds the optimal solution in $\mathcal O(m^2)$ time.
\end{proposition}

The classic bottleneck for approaches based on hindsight optimization is the (exponential, often) computational complexity of the associated solution. Yet, Proposition~\ref{prop:complexity_hindsight} shows that the particular problem structure allows us to drastically reduce the complexity of the recursive solution, which makes the Imitation Learning method described below particularly appealing in this scenario.\\

\begin{algorithm}[t]
\begin{algorithmic}[1]
\FUNCTION{ImitateExpert($\D$, $p$, $q$)}
\STATE Initialize $\Omega\leftarrow \emptyset$
\STATE Initialize $\pi\leftarrow\code{RandomPolicy}$
\FOR{$\ell  = 1,\ldots,p$}
\FOR{$i = 1,\ldots,q$}
\STATE Sample $\rho_i = (x_{i,1},\tau_{i,1}),\ldots,(x_{i,m_i},\tau_{i,m_i})$ from $\D$
\STATE Initialize current state $s\leftarrow s_0$
\FOR{$j=0,\ldots,{m_i}$}
\STATE Compute $\hat{a} \leftarrow \operatorname*{\arg\min}_{a\in \A}Q^{\rho_i}_j(s, a)$
\STATE Add instance-label pair $(s, \hat{a})$ to $\Omega$
\IF{$\ell=1$}
\STATE $s\leftarrow \delta(s, x_{j+1}, \hat{a})$
\ELSE
\STATE $s\leftarrow \delta(s, x_{j+1}, a\sim\pi(s))$
\ENDIF
\ENDFOR
\ENDFOR
\STATE Set $\pi\leftarrow\operatorname*{\arg\min}_{\pi'}\sum_{(s, a)\in \Omega} \code{loss}(\pi'(s), a)$
\ENDFOR
\STATE \textbf{return} $\pi$
\ENDFUNCTION
\end{algorithmic}
\caption{Imitation Learning}
\label{alg:dagger}
\end{algorithm}

\noindent\textbf{Imitation Learning.} In addition to the best sequence of actions, the above approach also computes the $Q^\rho$ values using which we can infer the best action from any possible state at any step. This enables us to apply existing imitation learning algorithms to our problem using the hindsight optimal action as the ``expert'' action at any state. We use the dataset aggregation algorithm introduced in \cite{ross2011reduction} for imitation learning.

Our imitation learning algorithm using hindsight optimization is outlined in Algorithm~\ref{alg:dagger}. It maintains a dataset $\Omega$ of labelled pairs $(s, a)$ indicating that the policy should choose action $a$ if the current state is $s$. It also maintains the current estimate of the best policy $\pi$ which is initialized to a policy that outputs random actions. In each iteration of the algorithm, $q$ sequences are drawn from $\D$. In order to draw sequences from $\D$ we choose a time limit $T_d$ and use the dataset $\Gamma$ to sample sequences in the same way as described in Section~\ref{sec:drl}. For each sequence of data points, the corresponding sequence of states are calculated using the policy $\pi$ (or the hindsight optimal actions if it is the first iteration). New labelled pairs are added to $\Omega$ by collecting all visited states with corresponding actions computed using the $Q^\rho$ values. Finally, at the end of each iteration, $\pi$ is updated using supervised learning on $\Omega$. The $\code{loss}$ function is chosen to be the cross-entropy loss.

\section{Shipping Consolidation: A Real World Example}\label{sec:exp}
We describe the real world shipping consolidation problem faced by a North American company who has one transportation hub in the United States. In 3 recent quarters, it processed nearly $5K$ orders with a total weight of $7M+$(kg) and around 800 different destination cities in the US. The maximum capacity of a full truck it used was a total weight of $L=22K$ (kg).

\begin{figure}[t]
\centering
\begin{tabular}{cc}
\includegraphics[width=0.4\linewidth]{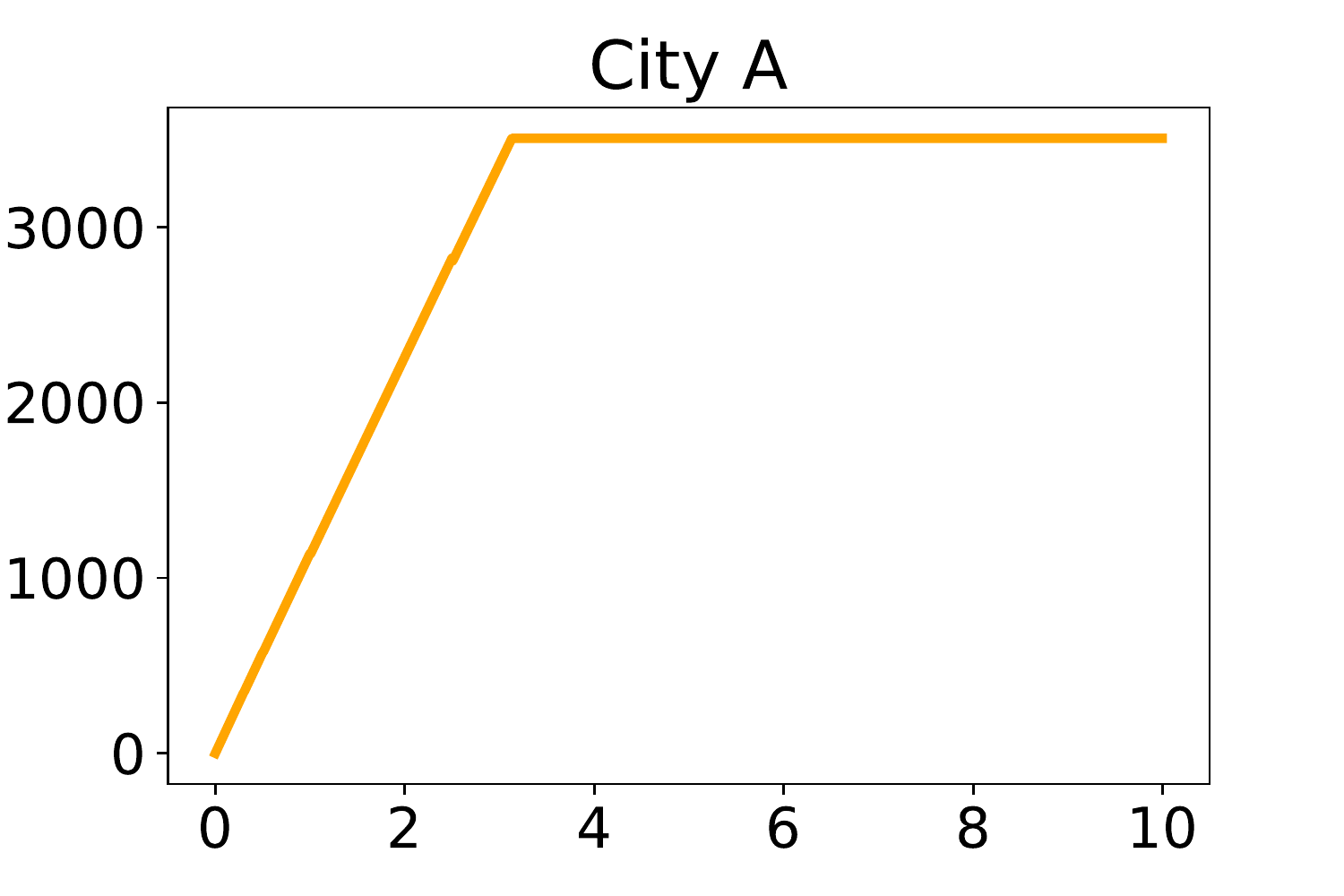} &
\includegraphics[width=0.4\linewidth]{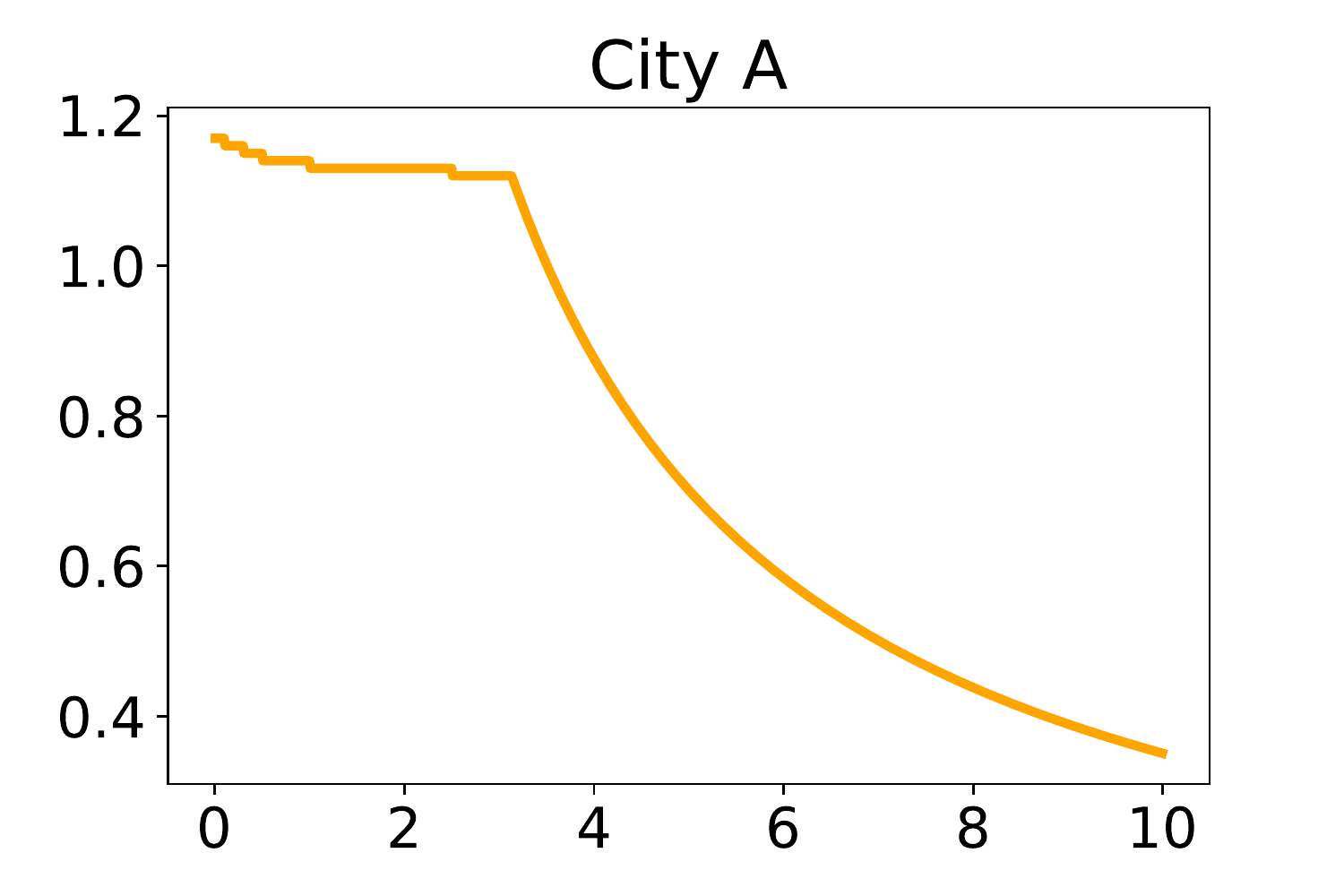}\\
(a) & (b) 
\end{tabular}
\caption{ $y$-axis is (a) total shipping cost or (b) shipping cost per Kg (in USD) and $x$-axis is the weight (in thousands of Kg) for some destination city A.}
\label{fig:cost}
\end{figure}

For any destination city, the shipping fee is a concave function of the weight similar to that shown in Figure~\ref{fig:cost} (a). Initially, the cost is increasing in the weight of the shipment (corresponding to the \emph{light-truck-load} regime, in logistics jargon) until the weight exceeds a certain threshold and the cost function flattens out (\emph{full-truck-load}---flat rate depending on the miles from hub to the destination). As a result, we can see that the shipping fee per unit weight decreases as the weight increases (Figure~\ref{fig:cost} (b)).

An important aspect of our dataset is that it consists of one order sequence per destination city. We also observed that the arrival rate of orders not only depends on the destination city but also changes with time. Due to these reasons, it is not possible to learn one policy for each city as it would not generalize well to future orders. Instead, we aim to learn a single neural network policy that works for all cities. Since the shipping cost $c_{\stop}$ depends on the destination city, we also include the maximum shipping cost $f(L)$ for the city in the state $s$. \\

\noindent \textbf{Implementation.} We now describe the details of how we implemented the three approaches for this specific problem.

\textit{Model-based approach.} We first discretize the range of possible weights $[0,L]$ into a finite grid $G$. Then, by treating each destination city \emph{separately}, we estimate the probability distribution of the order weights over the grid $G$ from previously observed input data via maximum-likelihood. This amounts to simply counting the number of order weights seen in each of the intervals in $G$. We estimate the expected inter-arrival time $\tau$ by simply averaging out past inter-arrival times. As explained in Algorithm \ref{alg:model}, such estimates are then used to construct and solve the continuous time MDP at each step. We remark that the optimal stopping strategy produced by \texttt{relaxed-stopping} sub-routine in Algorithm \ref{alg:model} is computed on the countable state space $G\times [0,\dots,I]$, where $G$ acts as the set of all possible truck load values and $I$ denotes the maximum possible number of items waiting to be shipped.

\textit{Deep reinforcement learning.} We use the available data to learn a single policy to be applied to \emph{all} the destination cities. Furthermore, the state is augmented with additional information to improve decision making; the input to the neural network policy is $(\ell, n, d, f(L), \hat{\tau}, \hat{x})$ where $d$ is the current total delay of orders held at the warehouse, $\hat{\tau}$ and $\hat{x}$ are the averages of inter-arrival times and order weights since the start of the episode. The maximum shipping cost $f(L)$, along with $\hat{\tau}$ and $\hat{x}$, serve as features that can be used to distinguish different cities whereas $d$ can be used to enforce hard constraints on the delay. We use two DRL algorithms to train policies, namely, DQN \citep{mnih2015human} and PPO \citep{schulman2017proximal}. We use the implementations of DQN and PPO found in StableBaselines \citep{stable-baselines}. 

\textit{Imitation Learning.} Similar to the DRL approach, we augment the state by adding the above features before storing it in the labelled dataset $\Omega$ in Algorithm~\ref{alg:dagger}. We also add an $\ell_2$-regularization loss to the supervised learning objective in order to avoid overfitting the dataset $\Omega$.

For both DRL and IL approaches, we used a neural network with two hidden layers and 32 nodes each to represent the policy. We employed the shipping data from the first two quarters for training and the last quarter for testing, respectively. \\

\setlength{\tabcolsep}{0pt}
\begin{figure*}[t]
    \centering
    \begin{tabular}{cccc}
    \vcentered{\includegraphics[width=0.29\linewidth]{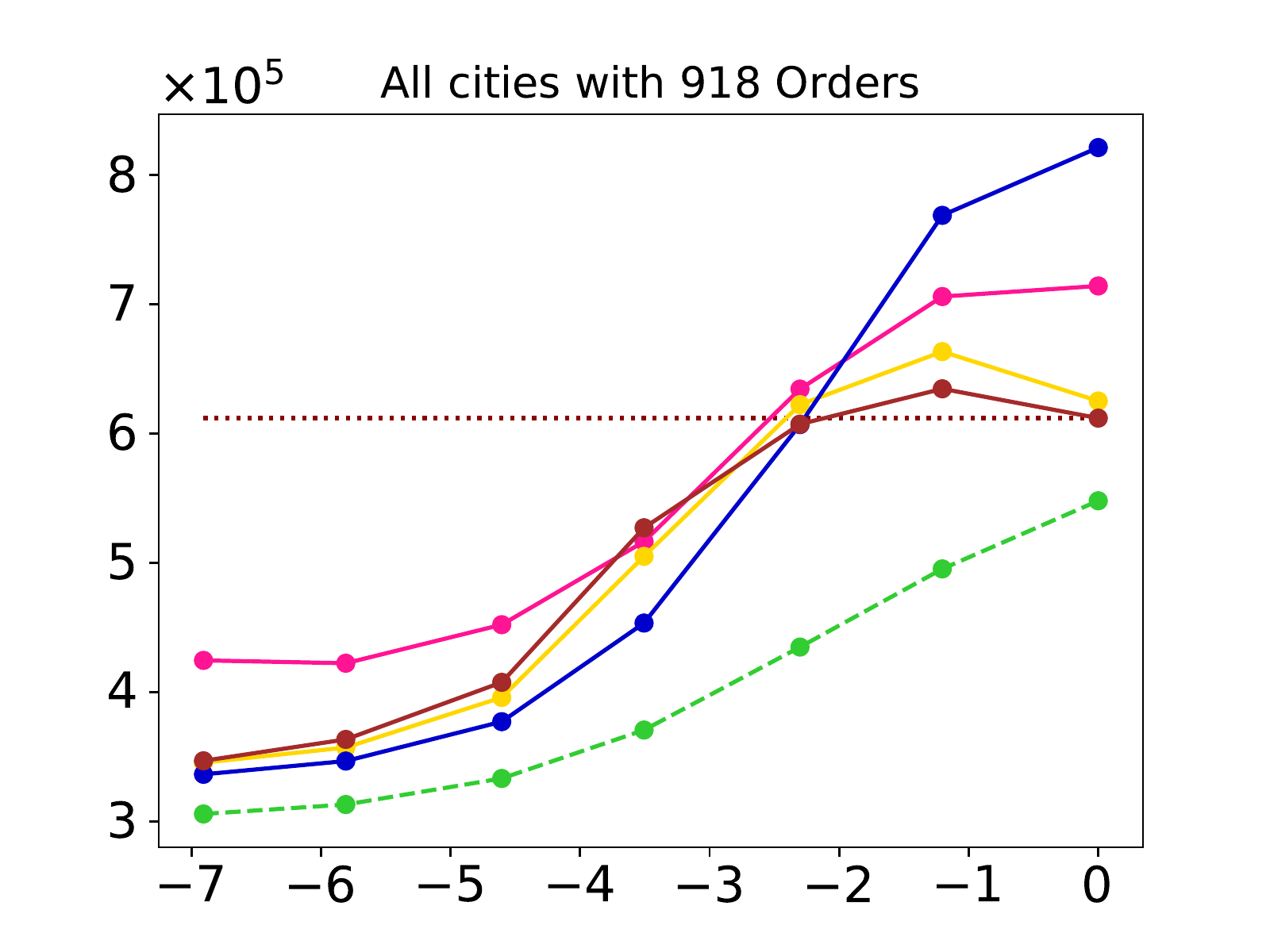}} &
    \vcentered{\includegraphics[width=0.29\linewidth]{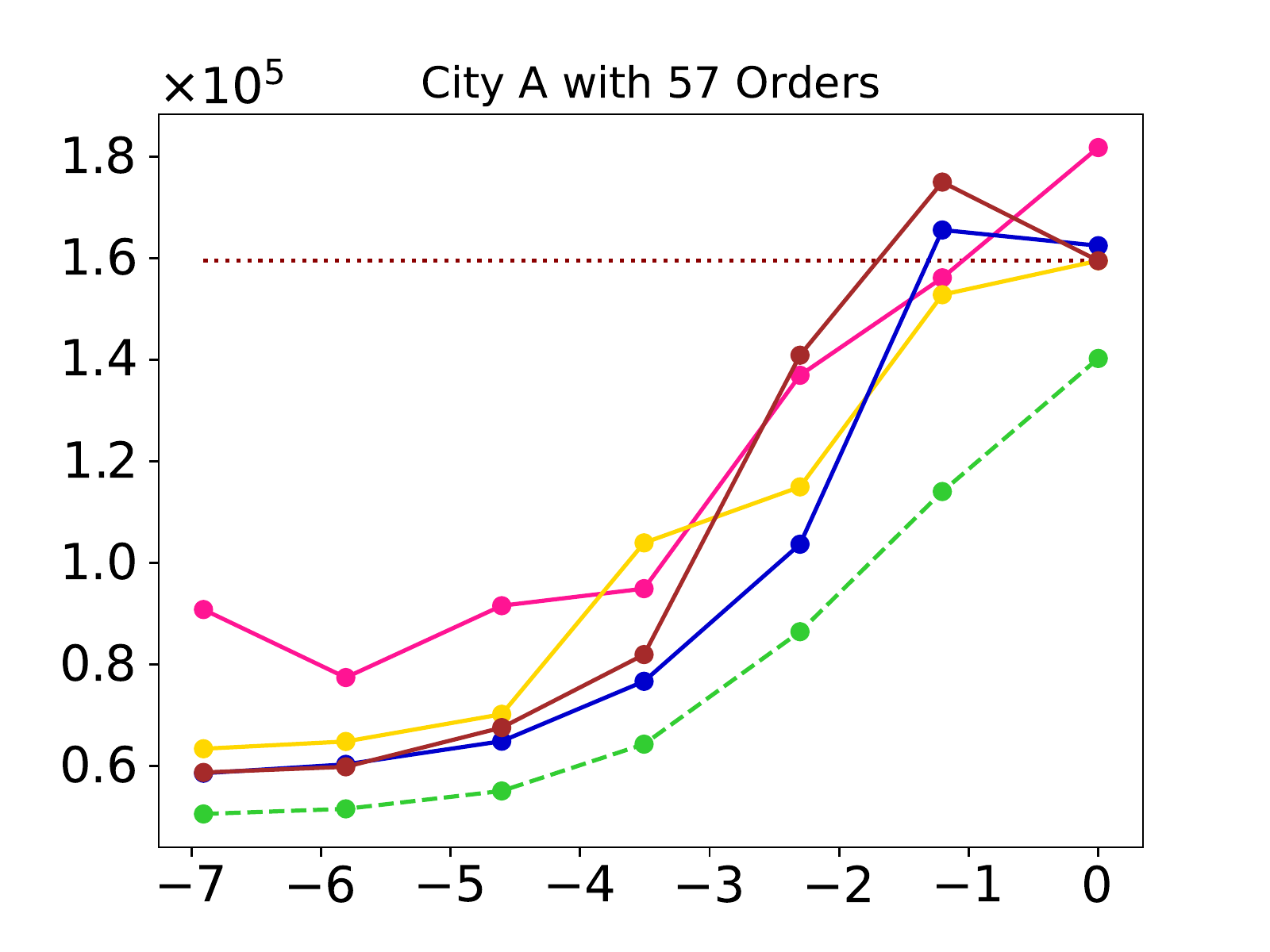}} &
    \vcentered{\includegraphics[width=0.29\linewidth]{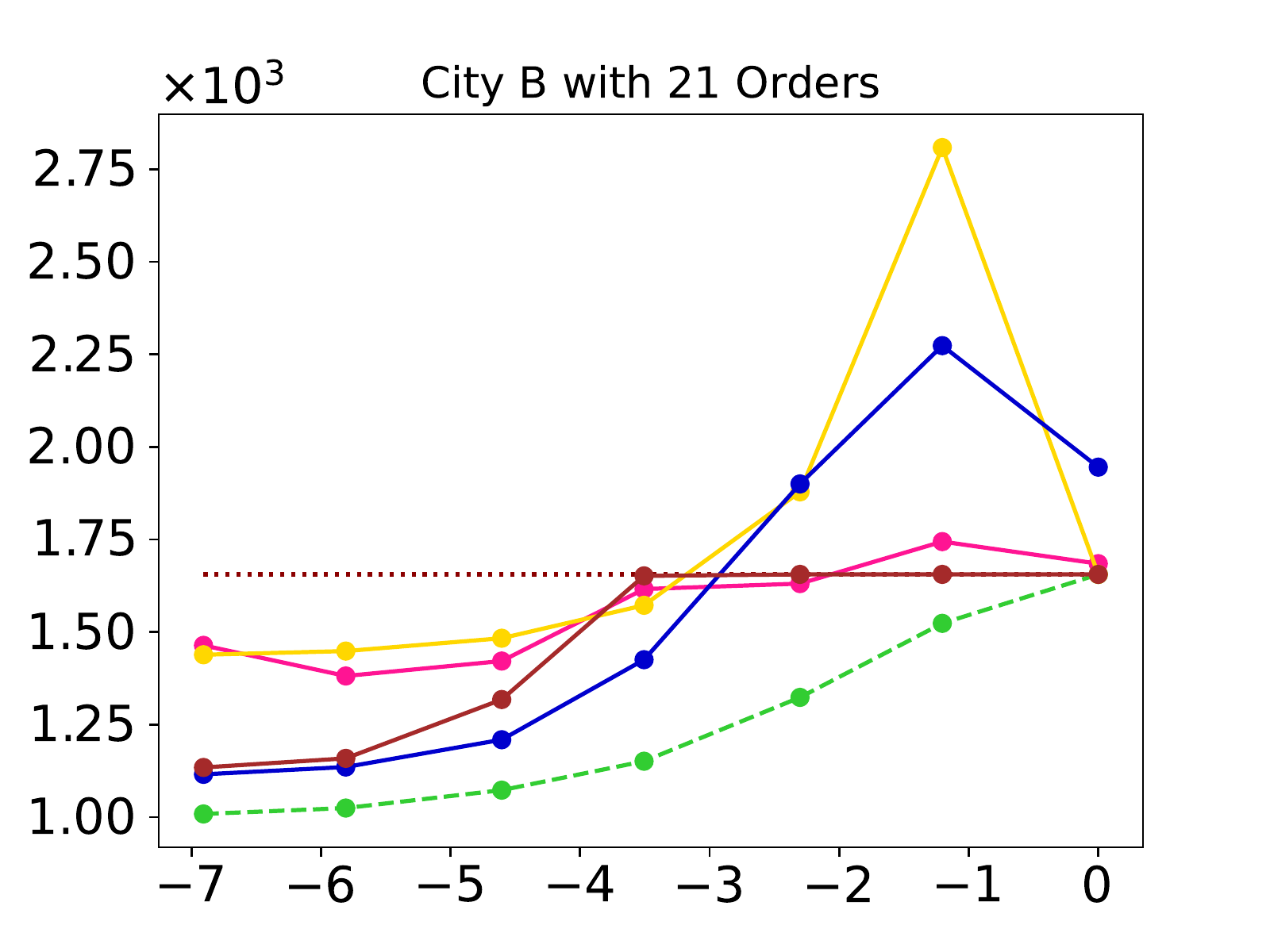}} &
    \vcentered{\includegraphics[width=0.12\linewidth]{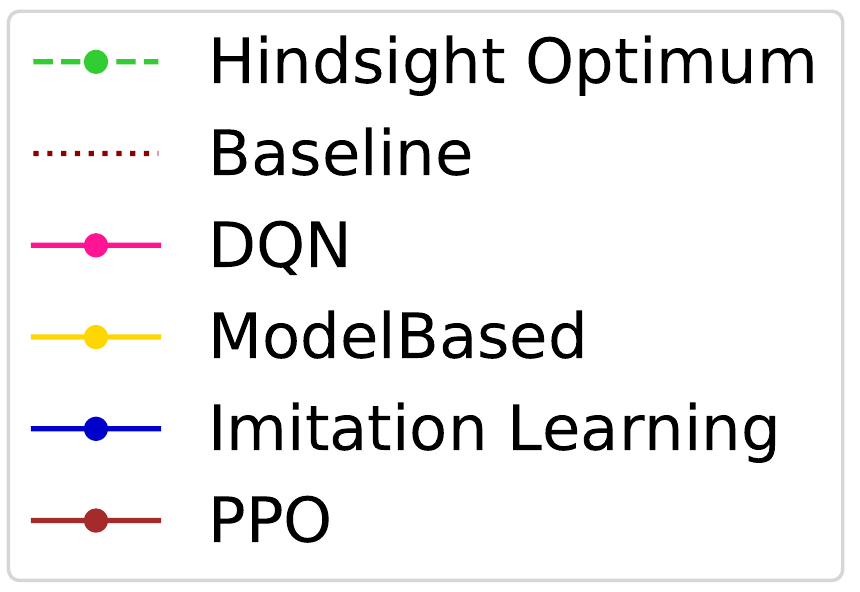}}\\
    (a) & (b) & (c) & 
    \end{tabular}
    \caption{Graphs showing the total cost (shipping cost and delay cost) in $y$-axis as a function of $\log(\alpha)$ ($x$-axis) for different approaches and destination cities. Results are averaged over 10 runs.}
    \label{fig:total_cost}
\end{figure*}

\setlength{\tabcolsep}{0pt}
\begin{figure*}[t]
    \centering
    \begin{tabular}{cccc}
    \vcentered{\includegraphics[width=0.29\linewidth]{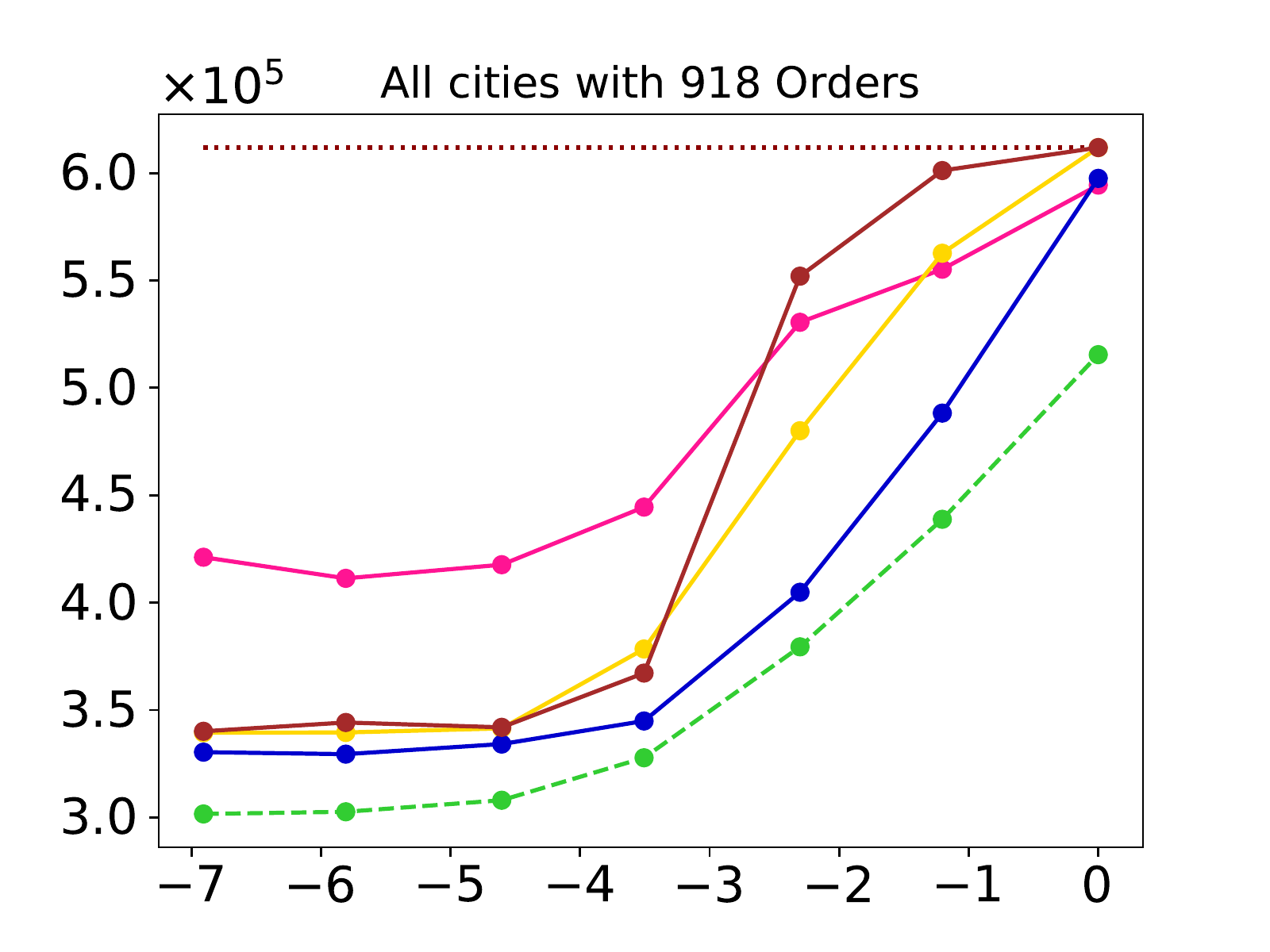}} &
    \vcentered{\includegraphics[width=0.29\linewidth]{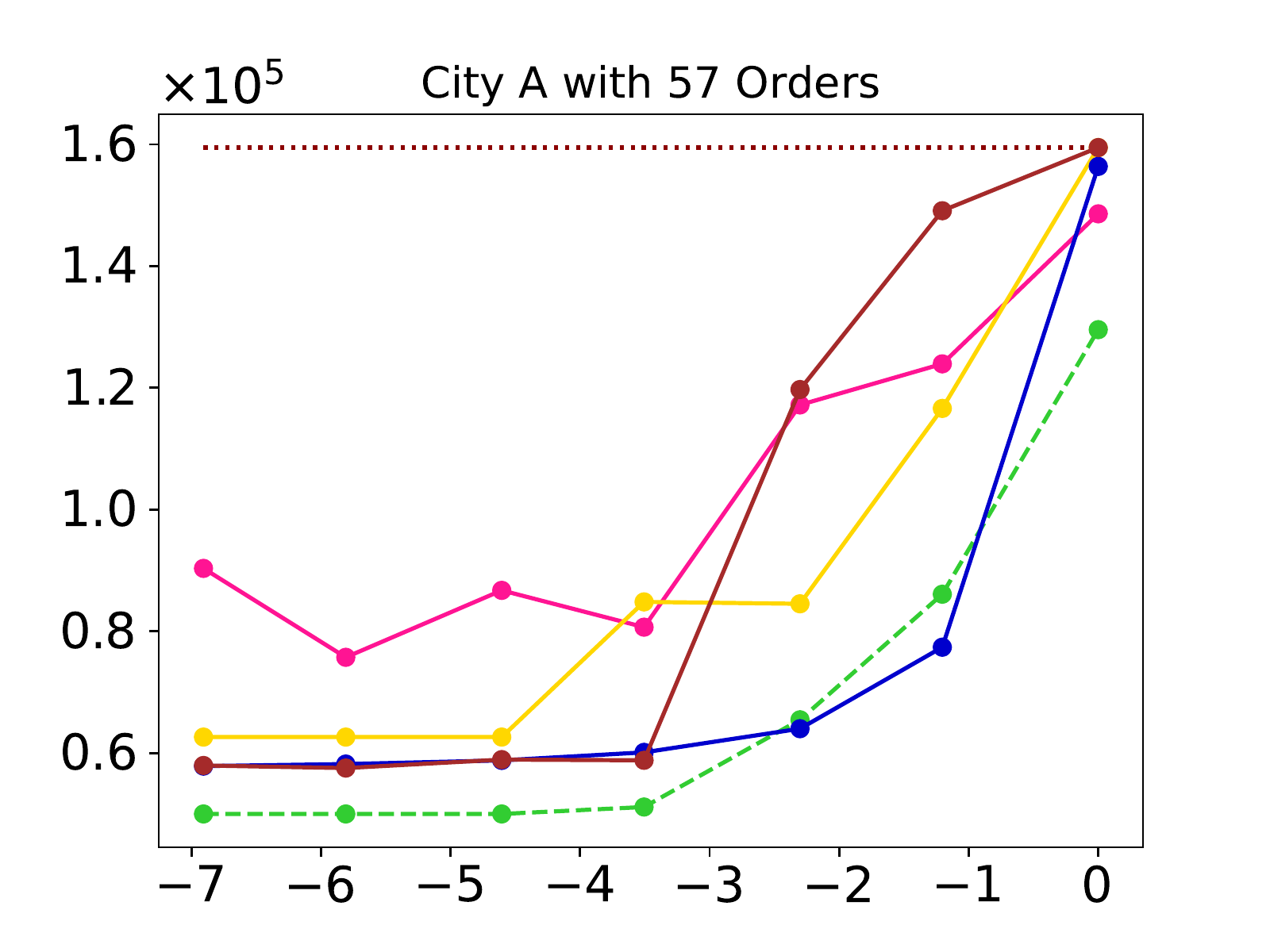}} &
    \vcentered{\includegraphics[width=0.29\linewidth]{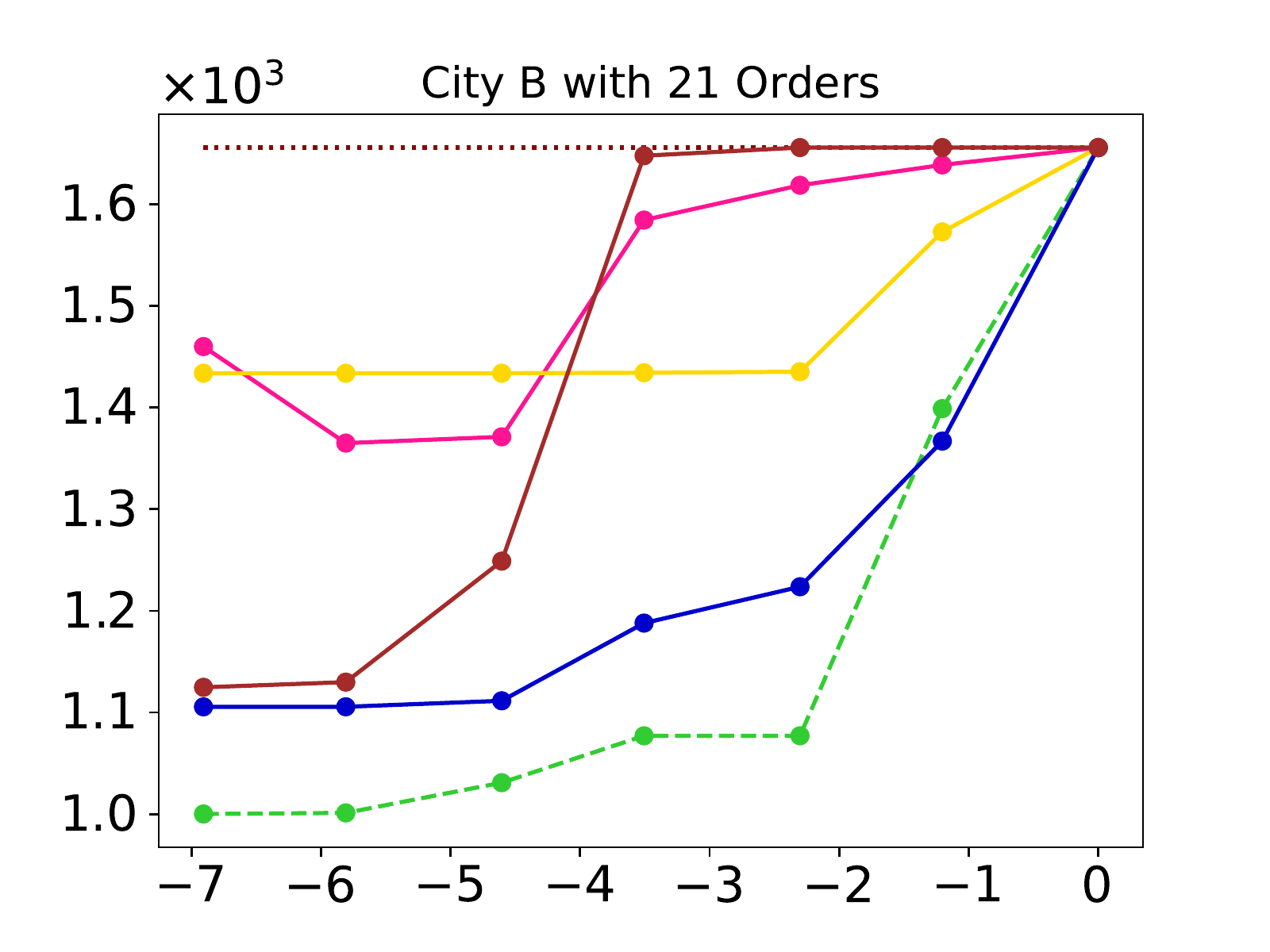}} &
    \vcentered{\includegraphics[width=0.12\linewidth]{plots/legend.pdf}}\\
    (a) & (b) & (c) & 
    \end{tabular}
    \caption{Graphs showing the shipping cost (in USD) in $y$-axis as a function of $\log(\alpha)$ ($x$-axis) for different approaches and destination cities. Results are averaged over 10 runs.}
    \label{fig:shipping_cost}
\end{figure*}

\setlength{\tabcolsep}{0pt}
\begin{figure*}[t]
    \centering
    \begin{tabular}{cccc}
    \vcentered{\includegraphics[width=0.29\linewidth]{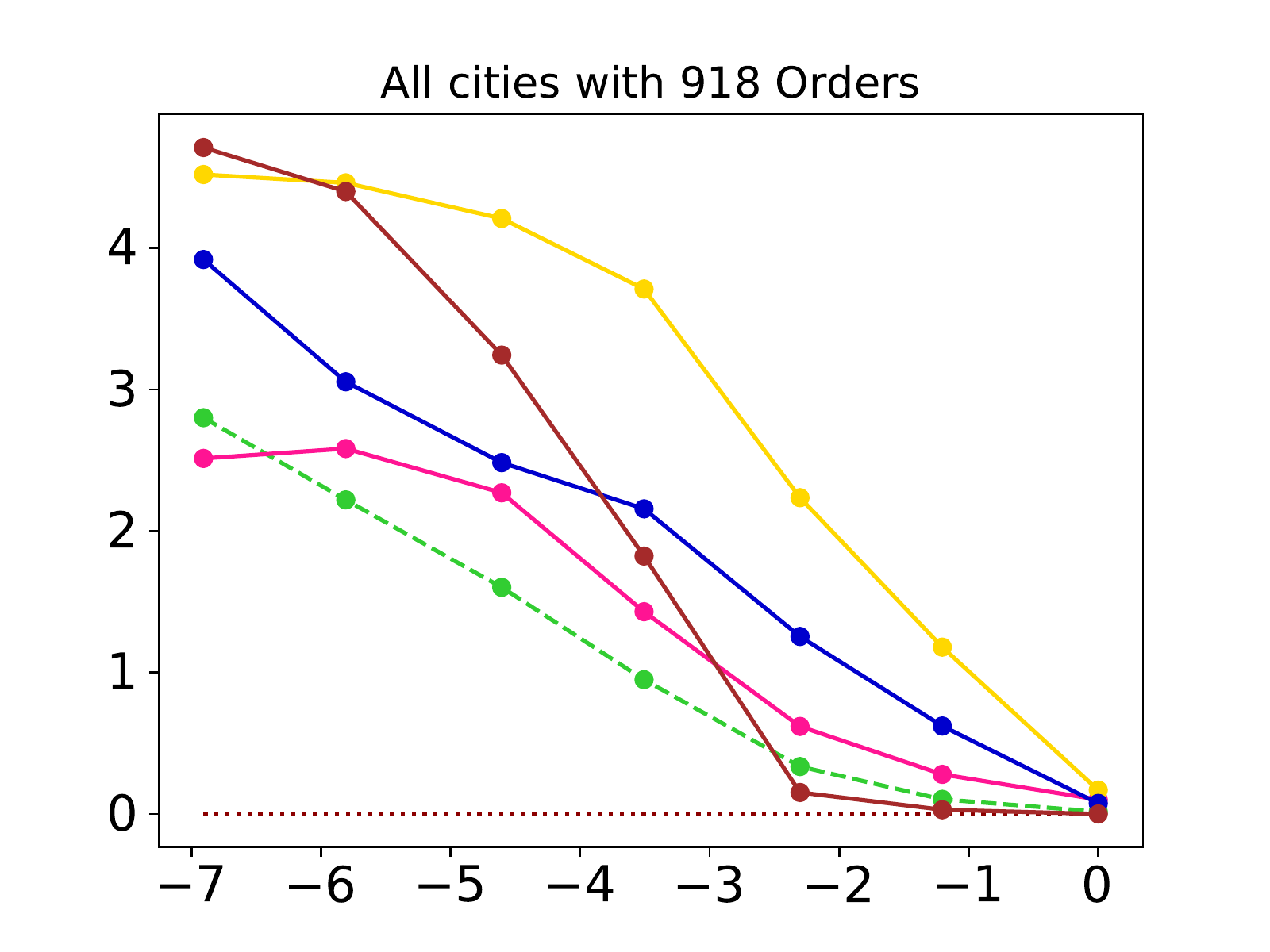}} &
    \vcentered{\includegraphics[width=0.29\linewidth]{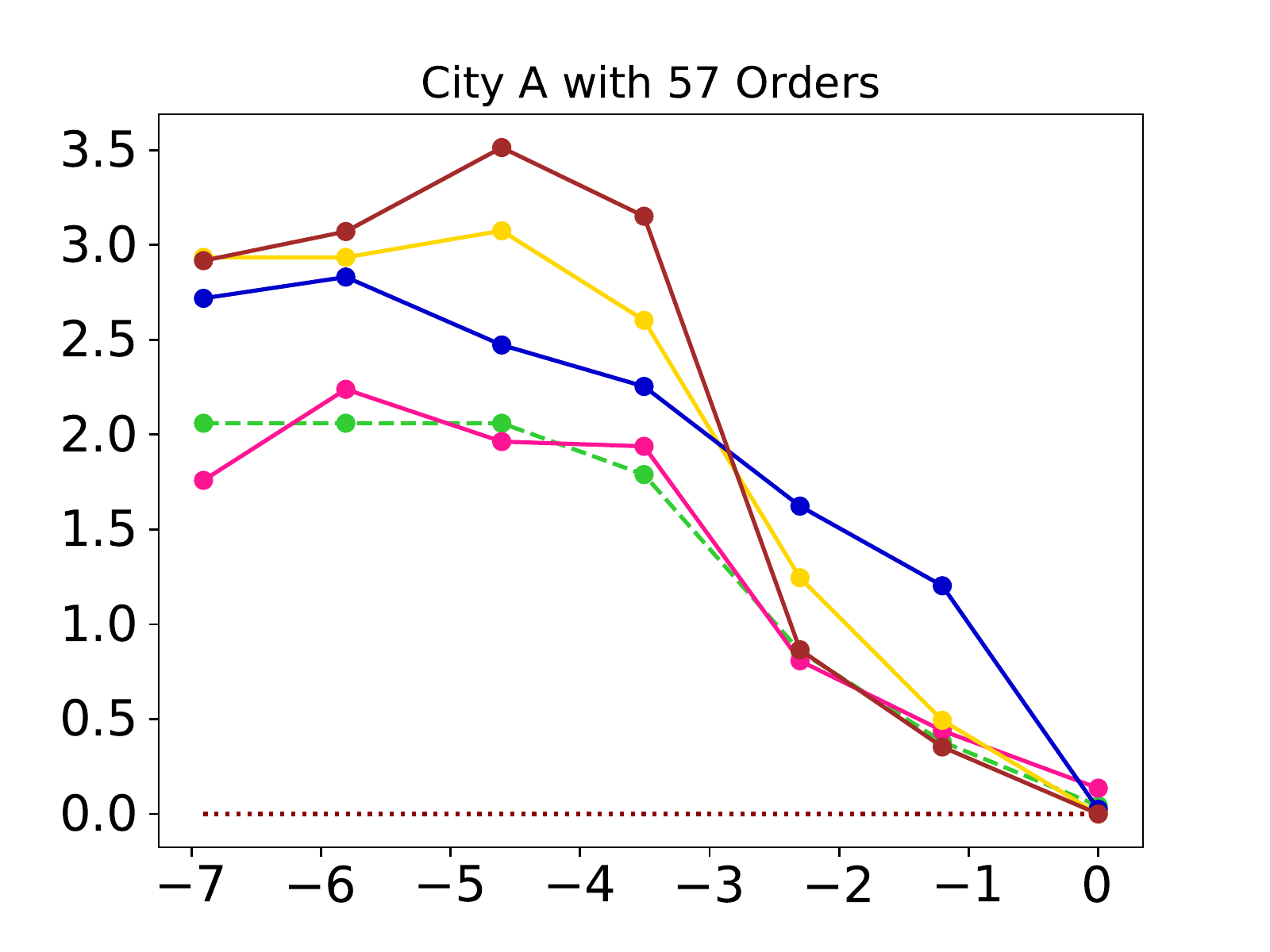}} &
    \vcentered{\includegraphics[width=0.29\linewidth]{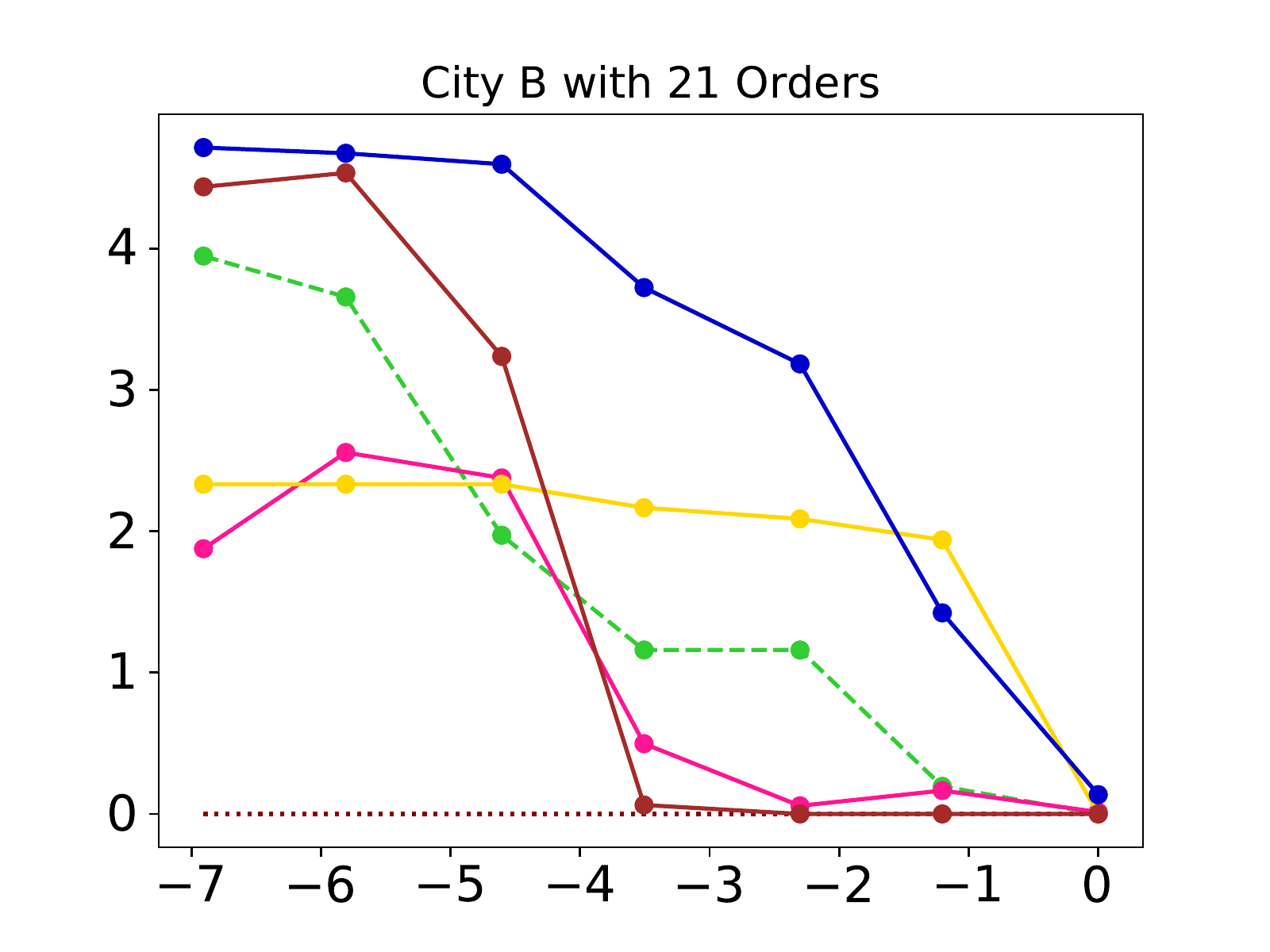}} &
    \vcentered{\includegraphics[width=0.12\linewidth]{plots/legend.pdf}}\\
    (a) & (b) & (c) & 
    \end{tabular}
    \caption{Graphs showing the delay per order (in days) in $y$-axis as a function of $\log(\alpha)$ ($x$-axis) for different approaches and destination cities. Results are averaged over 10 runs.}
    \label{fig:delay_cost}
\end{figure*}

\noindent \textbf{Results.} The results are shown in Figures~\ref{fig:total_cost}, \ref{fig:shipping_cost} and \ref{fig:delay_cost}. We compare the performance of the different approaches for different values of $\alpha$, which governs the tradeoff between shipping cost and delay. Figure~\ref{fig:total_cost} shows the total cost over one quarter (corresponding to the test data) as a function of $\log(\alpha)$. Figures~\ref{fig:shipping_cost} and \ref{fig:delay_cost} show the total shipping cost and the delay per order as a function of $\log(\alpha)$, respectively. 
In each figure, the result for all cities (sum of costs for all cities) is labelled (a) and results for city A and city B are labelled (b) and (c), respectively. City A has a relatively larger number of orders as compared to city B in the test quarter. The baseline strategy is a policy that incurs no delay and consolidates a shipping as soon as one order arrives. 
Hindsight optimum denotes the solution given by offline hindsight optimization performed on the \emph{whole} test data sequences; clearly, this is not a feasible online strategy since it peeks into the future to make decisions.\\

\noindent \textit{Model-based approach.} It performs reasonably well while requiring very limited training data, since it only requires to predict the order weight distribution and inter-order arrivals, and the MDP solution turns out to be robust to prediction errors. On the other hand, the complexity at run-time is considerably higher than the other two techniques, since it solves the \texttt{MDP} routine of Algorithm \ref{alg:model} every few steps, while NN-based approaches just require to perform an NN inference given the current state. Moreover, its performance greatly depends on the ability to predict future arrivals. In its current implementation, it struggles to adapt to changes in the input data distribution since the prediction is performed via simple averages. \\

\noindent \textit{Deep RL.} We observe that DQN is somewhat unstable and performs poorly as compared to PPO for this task, likely because it is harder to learn a single Q-network for all destination cities. PPO achieves good performance and is comparable to the model based approach in terms of the total cost. We remark that DRL can be seen as a black box tool here and is not explicitly taking advantage of the structure in the problem. Nonetheless, we are able to learn good stopping strategies using this approach.\\

\noindent \textit{Imitation learning.} It achieves the minimum total cost among all tested approaches for small values of $\alpha$ and is closest to the hindsight offline optimum, as shown in Figure \ref{fig:total_cost}. We remark that its complexity at run-time is considerably lower than the model-based approach, since it does not require to solve the MDP at each step. Moreover, its training complexity is greatly reduced via the efficient DP solution to the hindsight optimization problem. The only hitch arises when $\alpha$ increases, since IL shows the tendency of waiting too long. Yet, in this case the shipping delay term becomes preponderant and the optimal strategy should tend to the baseline solution that incurs no delay. In contrast, PPO learns this behavior more efficiently.\\



\setlength{\tabcolsep}{2pt}
\begin{figure}[t]
    \centering
    \begin{tabular}{cc}
    \vcentered{\includegraphics[width=0.4\linewidth]{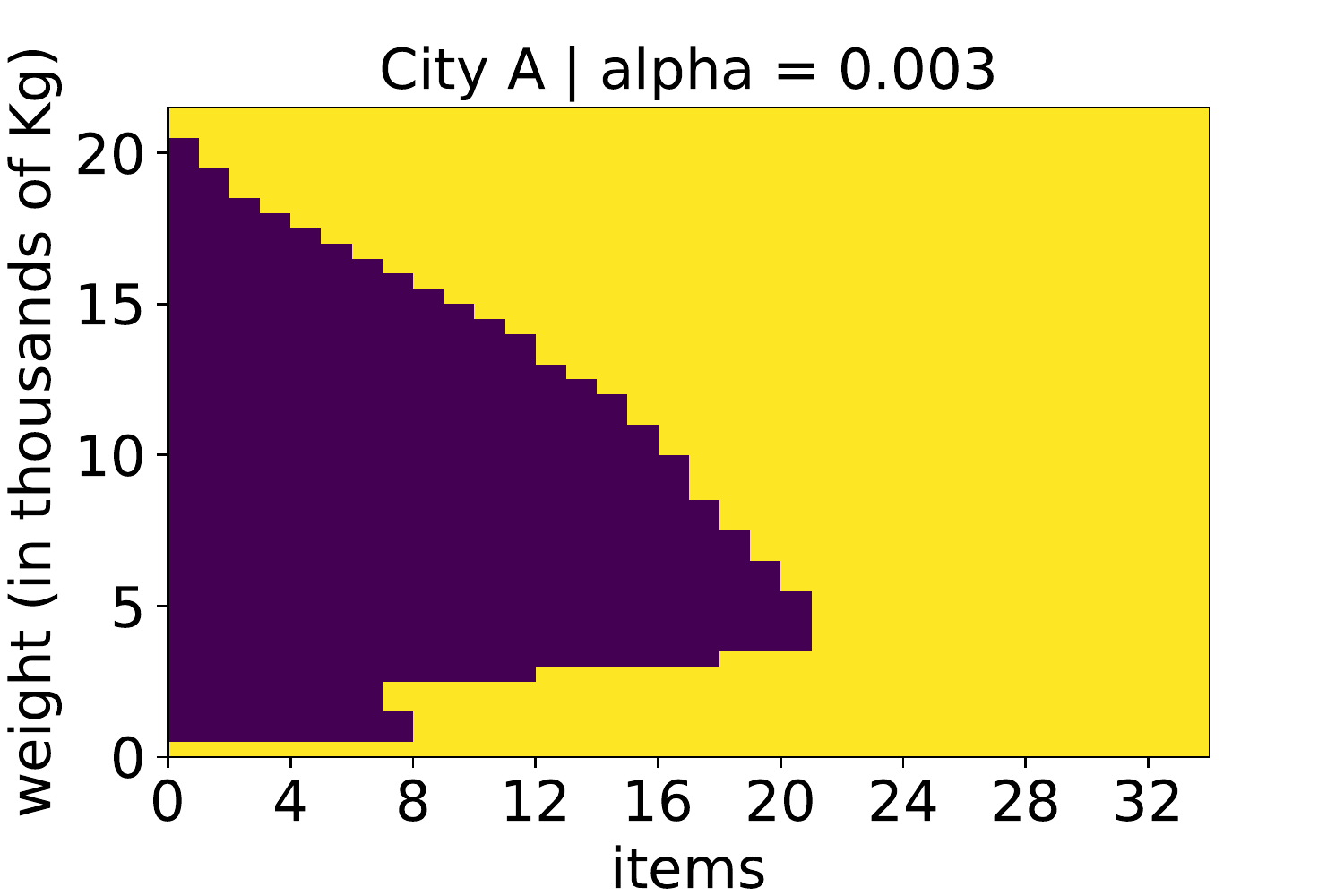}} &
    \vcentered{\includegraphics[width=0.4\linewidth]{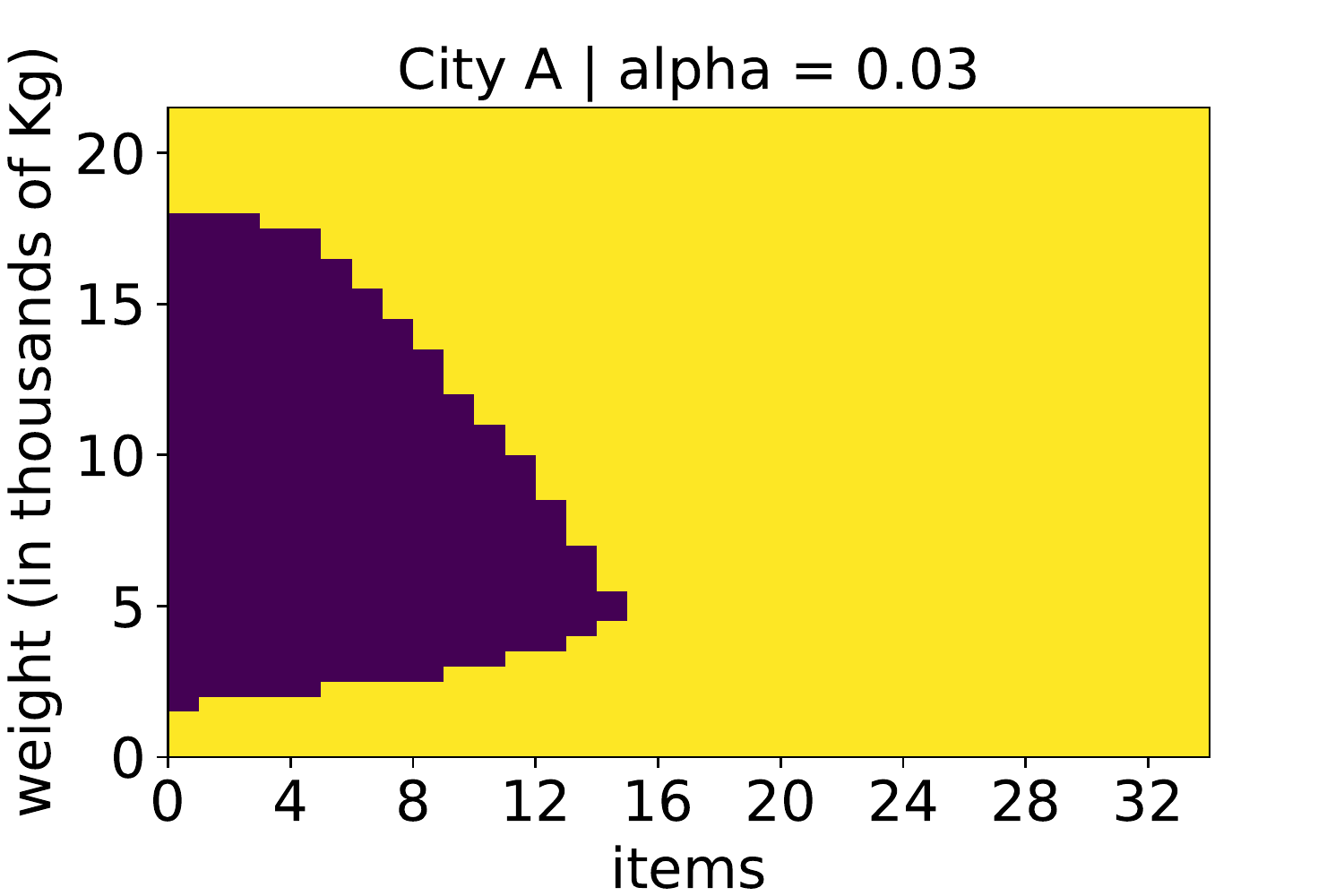}}\\
    (a) & (b)
    \end{tabular}
    \caption{Visualization of the neural network policies learned using imitation learning for different values of $\alpha$. Yellow and violet regions corresponds to \stop{} and \wait{} actions, respectively.}
    \label{fig:nn_policy}
\end{figure}

\setlength{\tabcolsep}{2pt}
\begin{figure}[t]
    \centering
    \begin{tabular}{cc}
    \vcentered{\includegraphics[width=0.4\linewidth]{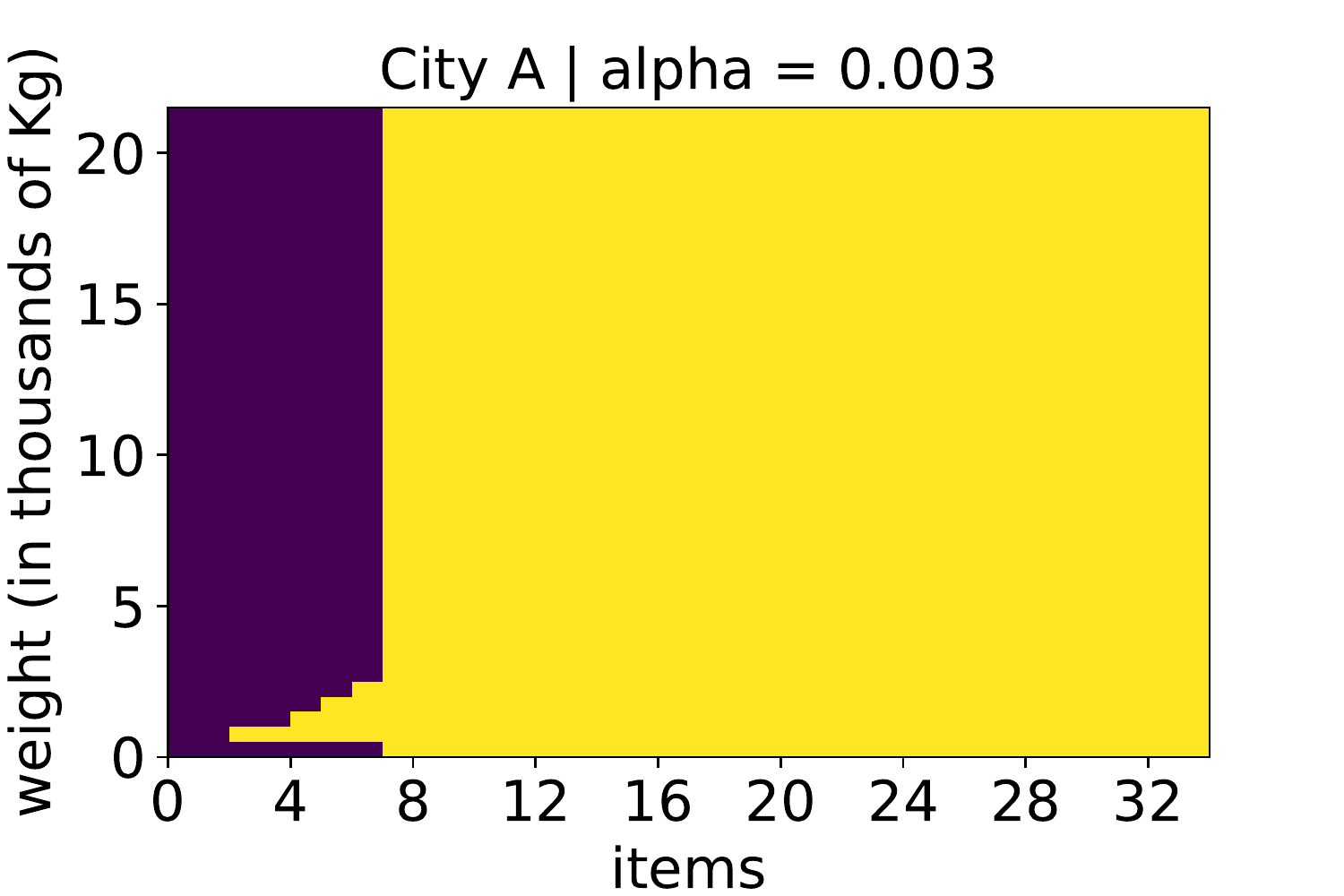}} &
    \vcentered{\includegraphics[width=0.4\linewidth]{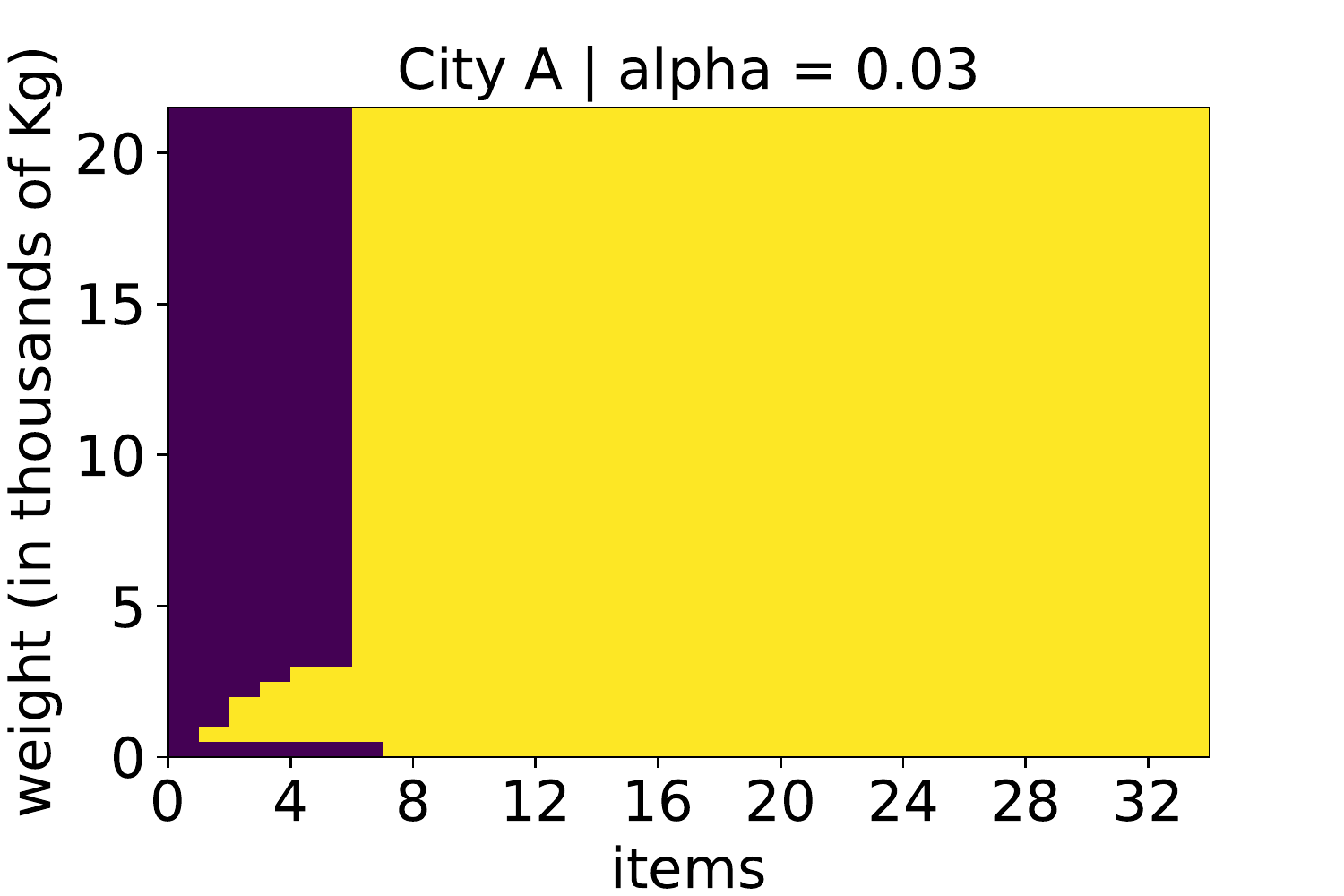}}\\
    (a) & (b)
    \end{tabular}
    \caption{Visualization of stopping strategies computed by the model-based algorithm for different values of $\alpha$. Yellow and violet regions corresponds to \stop{} and \wait{} actions, respectively.}
    \label{fig:policy_model}
\end{figure}

\noindent \textbf{Policy Visualization.} To gather more intuitions, it is useful to visualize the policy as a function of the weight and number of items as shown in Figure~\ref{fig:nn_policy}. The yellow region is where the policy selects the $\stop$ (i.e., consolidate) action while the violet region corresponds to the $\wait$ action. As expected, the yellow region broadens as $\alpha$ increases, since reducing the delay is of greater importance. To produce Figure~\ref{fig:nn_policy} we kept the remaining inputs to the neural networks as fixed, and we computed them based on test data corresponding to a specific city destination. 

{Similarly, we also plot stopping strategies computed by the model-based approach in Figure~\ref{fig:policy_model}. These policies are optimal for the model estimated from the test data. Since the model itself is an estimate of a changing process, it’s not necessarily the case that an optimal policy based on the model is an optimal policy for the actual data. As a result, the neural network policies differ from these policies on some inputs. }

\section{Conclusions}\label{sec:concl}
We studied different learning approaches for online decision making in the context of regenerative stopping problems, where the system resets after a \stop{} action is chosen. Regenerative problems arise naturally in real-world scenarios such as shipping consolidation in logistics, where one decides when to consolidate a truck to strike a balance between shipping cost and delay.

Model-based solutions rely on an efficient technique by \cite{miller1981countable} to solve the underlying MDP. Yet, (i) their performance depend greatly on the ability to predict future inputs and (ii) show high run-time complexity, since a new MDP has to be solved whenever the prediction is updated. On the other hand, deep learning approaches (i) can adapt naturally to changes in input distribution as they directly learn a policy from historical data, and (ii) only require a NN inference at run-time.


%
%
%


\bibliographystyle{abbrvnat}
\bibliography{main}


\end{document}